\documentclass[twoside,11pt]{article}
\usepackage{jmlr2e_without_decorations}
\usepackage{graphicx} 

\usepackage{xcolor,graphicx,amsmath,amssymb,bm}
\usepackage[parfill]{parskip}

\usepackage{svg}

\firstpageno{1}
\begin{document}
\title{Spatial Graph Coarsening: Weather and Weekday Prediction with London's Bike-Sharing Service using GNN}

\author{\name Yuta Sato \email \href{mailto:y.sato4@lse.ac.uk}{y.sato4@lse.ac.uk} \\ \addr{Department of Geography and Environment \\ London School of Economics and Political Science}
\AND
\name Pak Hei Lam  \email \href{mailto:p.h.lam1@lse.ac.uk}{p.h.lam1@lse.ac.uk} \\ \addr{Department of Statistics \\ London School of Economics and Political Science}
\AND
\name Shruti Gupta  \email \href{mailto:s.gupta121@lse.ac.uk}{s.gupta121@lse.ac.uk} \\ \addr{Department of Statistics \\ London School of Economics and Political Science}
\AND
\name Fareesah Hussain  \email \href{mailto:a.i.hussain@lse.ac.uk
}{a.i.hussain@lse.ac.uk} \\ \addr{Department of Statistics \\ London School of Economics and Political Science}
}

\maketitle

\begin{abstract}%
    This study introduced the use of Graph Neural Network (GNN) for predicting the weather and weekday of a day in London, from the dataset of Santander Cycles bike-sharing system as a graph classification task. The proposed GNN models newly introduced (i) a concatenation operator of graph features with trained node embeddings and (ii) a graph coarsening operator based on geographical contiguity, namely \textit{Spatial Graph Coarsening}. With the node features of land-use characteristics and number of households around the bike stations and graph features of temperatures in the city, our proposed models outperformed the baseline model in cross-entropy loss and accuracy of the validation dataset.
\end{abstract}

\begin{keywords} 
Graph Neural Network, Graph Classification, Graph Coarsening, Bike Sharing Service, Spatial Graph Coarsening
\end{keywords}

\section{Introduction}
    Urban transportation systems are essential for modern cities as they provide mobility to inhabitants, while significantly impacting the environmental, economic, and social aspects of cities. A form of transportation that is growing in popularity in recent years is biking. Bike-sharing services have become prevalent in cities worldwide and have emerged as a valuable tool for promoting sustainability as well as a healthier lifestyle. These services can also be viewed as socio-economic data sources that could provide useful insights into human activities in urbanised areas, with potential applications in fields such as Urban Economics and Social Network Analysis.

    In parallel, since the Graph Neural Network (GNN) became prevalent for graph representation learning, a many algorithms have been invented, such as Graph Convolution Network (GCN; \citealt{kipf2017semisupervised}). Since GNN-based models are permutation-invariant and free from the degree of the nodes for the neural network architecture, they have been applied to a variety of domains, such as molecular property prediction \citep{wang2023graph}; and social network recommendation \citep{fan2019graph}. In particular, demand prediction for bike-sharing and ride-sourcing services has been one of the notable topics of GNN-based model applications, for example in Beijing \citep{lin2018} and in New York \citep{ke2021}. However, previous studies have focused only on the node regression of bike usage per station and edge regression of travel record counts, and graph prediction is still understudied. Moreover, the GNN-based models in the previous studies lacked the consideration of the geographical contiguity of bike stations as nodes, which is due to the graph representation of the dataset only as the relationship of travel records, not as the spatial proximity. A drawback could be the failure to capture the homophily of geographical neighbours between nodes in the GNN-based model. Lastly, any GNN-based prediction tasks has not been investigated yet on London’s bike-sharing services.

    In this study, we conduct a prediction of the weather and weekday in London from the usage of Santander Cycles \citep{tfl}, in order to demonstrate the importance of understanding the characteristics of a city. We use GNN to predict the weather and weekday of daily graphs as graph embeddings, with mapped land-use characteristics \citep{fleischmann_arribas-bel_2022} and number of households in the neighbours by demographics \citep{census_2021_results_2022} as node features. Our main contributions are as follows (i) the introduction of a concatenation operator of graph features with trained node embeddings obtained after \textit{Global Mean Pooling}; and (ii) the introduction of graph coarsening operator based on geographical contiguity, namely \textit{Spatial Graph Coarsening}. These new proposals shall expand the employability of bike-sharing dataset for a better understanding of cities and humans living there. 

\section{Problem Formulation}
    \subsection{Graph Neural Networks}
    Let $\mathcal{G} = (\mathcal{V}, \mathcal{E})$ be a graph of the bike-sharing service dataset in a graph representation, where $\mathcal{V}$ and $\mathcal{E}$ denotes the bike stations used as nodes and travel records as edges on the day, and note that $\rvert \mathcal{V} \lvert$ and $\rvert \mathcal{E} \lvert$ stand for the number of nodes and edges of graph $\mathcal{G}$, respectively. Given $\mathbf{A} \in \mathbb{R}^{\lvert \mathcal{V} \rvert \times \lvert \mathcal{V} \rvert}$, $\mathbf{H}^{(k)} \in \mathbb{R}^{\lvert \mathcal{V} \rvert \times d^{(k)}}$, $\mathbf{W}^{(k)} \in \mathbb{R}^{d^{(k+1)} \times d^{(k)}}$, and $\mathbf{B}^{(k)} \in \mathbb{R}^{\lvert \mathcal{V} \rvert \times d^{(k+1)}}$which denote adjacency matrix of $\mathcal{G}$, node embeddings matrix of dimension $d^{(k)}$, and weight matrix and bias matrix at layer $k$, respectively, the node embeddings matrix at layer $k+1$ as a result of message passing of a GNN model is written as follows: $$\mathbf{H}^{(k+1)} = \sigma \left((\mathbf{I} + \mathbf{A}) \mathbf{H}^{(k)} (\mathbf{W}^{(k)})^{\top} + \mathbf{B}^{(k)}\right)$$
    where $\mathbf{I} \in \mathbb{R}^{\lvert \mathcal{V} \rvert  \times \lvert \mathcal{V} \rvert}$ is an identity matrix and $\sigma(\cdot)$ is an activation function. As our assumption, the weight matrix $\mathbf{W}^{(k)}$ can be shared between neighbours and self-nodes, since bike-sharing services are not limited to combination of different nodes (i.e. one can use a bike within Hyde Park). Regarding the adjacency matrix, we additionally use $\mathbf{A}_{\text{weighted}}$ as a weighted one, such that $[\mathbf{A}_{\text{weighted}}]_{ij} = n_{ij}$, where $n_{ij}$ denote the total count of travel records on a day between bike station $i$ and $j$. Moreover, $\mathbf{A}$ and $\mathbf{A}_{\text{weighted}}$ are used in the message passing after symmetrical normalisation as a process of \textit{Graph Convolution} \citep{kipf2017semisupervised}
     to balance out the influence of nodes, such that
    \begin{align*}
        \tilde{\mathbf{A}} &= \mathbf{D}^{-\frac{1}{2}} \mathbf{A} \mathbf{D}^{-\frac{1}{2}} \\
        \tilde{\mathbf{A}}_{\text{weighted}} &= \mathbf{D}_{\text{weighted}}^{-\frac{1}{2}} \mathbf{A}_{\text{weighted}}\mathbf{D}_{\text{weighted}}^{-\frac{1}{2}}
    \end{align*}
    where $\mathbf{D}$ and $\mathbf{D}_{\text{weighted}} \in \mathbb{R}^{\rvert \mathcal{V} \lvert \times \rvert \mathcal{V} \lvert}$ denote degree matrices of $\mathbf{A}$ and $\mathbf{A}_{\text{weighted}}$. A degree matrix is a diagonal matrix where the entires represent the degree of a vertex, i.e. the number of times an edge terminates at that vertex.


    
    
    \subsection{Global Mean Pooling and Graph Features Concatenation} \label{globalmeanpool}
    In GNN architectures, graph pooling is a popular operation to produce summarised information of graphs for graph prediction tasks \citep{Hamilton}. This study commonly employs the \textit{Global Mean Pooling} operator to obtain graph embeddings for following fully-connected layers. After \textit{Global Mean Pooling}, a graph embedding with $d_{\mathrm{graph}}$ components $\mathbf{z}_{\mathcal{G}} \in \mathbb{R}^{d_{\mathrm{graph}}}$ is given by $\mathbf{z}_{\mathcal{G}} = \left(\sum_{v \in \mathcal{V}}\mathbf{z}_u \right) / \lvert \mathcal{V} \rvert$, where $\mathbf{z}_u \in \mathbb{R}^{d_{\mathrm{node}}}$ is a vector of node embedding for node $u$ with $d_{\mathrm{node}}$ components after graph convolutions. In addition to the obtained graph embeddings, we utilise graph features of temperature as a representation of seasonality. Let $\mathbf{x}_{\mathcal{G}} \in \mathbb{R}^{3}$ be a vector of graph features of mean, maximum and minimum daily temperatures. Then, we concatenate the graph embeddings and the graph features of temperatures to obtain $\mathbf{z}_{\mathcal{G}} \oplus \mathbf{x}_{\mathcal{G}} = \tilde{\mathbf{z}}_\mathcal{G} \in {\mathbb{R}^{d_{\mathrm{graph}}+3}}$, where $\oplus$ denotes concatenation of two vectors.

    
    \subsection{Spatial Graph Coarsening}
    \subsubsection{KNN graph}    
    \begin{figure}[!h]
        \centering
        \includegraphics[width=\textwidth]{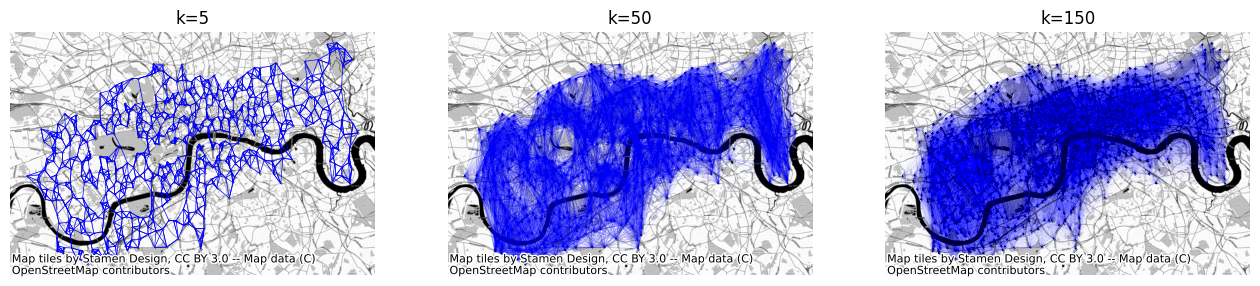}
        \caption{KNN graph on bike stations of Santander Cycles in London. Edges were generated based on $k \in \{5,50,150\}$ nearest neighbours of geographic contiguity.}
        \label{Fig:knngraph}
    \end{figure}
    Although the framework of GNN we introduced could capture the heterogeneity of weather and weekday, it lacks in considering the information of geographical contiguity between nodes. Thus, instead of directly pooling all the node embeddings into a single vector, we would like to propose spatial graph coarsening, which is based on clusters on $k$-Nearest Neighbour (KNN) graph of nodes by the geographical coordinates. By introducing spatial graph coarsening, the similarities among geographic neighbours could be aggregated by clusters. As shown in Figure \ref{Fig:knngraph}, $k$ as the number of neighbours determines the connectivity in the KNN graph. Let $\mathcal{G}_{\mathrm{KNN}}^{(k)} = (\mathcal{V}, \mathcal{E}_{\mathrm{KNN}}^{(k)})$ a KNN graph based on $k$ nearest neighbours on the basis of Euclidian distance of the geographic coordinates, the component of its adjacency matrix $\mathbf{A}_{\mathrm{KNN}}^{(k)} \in \mathbb{R}^{{\lvert \mathcal{V}} \rvert \times \lvert \mathcal{V} \rvert}$ is written as follows:
    $$
    [\mathbf{A}_{\mathrm{KNN}}^{(k)}]_{ij} = 
      \begin{cases}
        1   & \text{if $j \in \mathcal{N}^{(k)}(i)$ $\forall i, j \in \mathcal{V}$} \\
        0   & \text{otherwise.}
      \end{cases}
    $$
    where $i, j \in \mathcal{V}$ denote the nodes in the KNN graph, which is invariant to the value of $k$, and $\mathcal{N}^{(k)}(i)$ stands for the $k$ nearest neighbours of node $i$.
    
    \subsubsection{Spectral Clustering}
    To construct clusters on the KNN graph $\mathcal{G}_{\mathrm{KNN}}^{(k)}$, we employ Spectral Clustering to identify strongly connected communities of nodes in graphs, by minimising Equation \ref{eq:ncut},
    \begin{align} \label{eq:ncut}
        \text{NCut}(\mathcal{A}_1, \dots, \mathcal{A}_K) = \frac{1}{2} \frac{\sum_{k=1}^{K} \lvert (u,v) \in \mathcal{E}: \; u \in \mathcal{A}_k, v \in \bar{\mathcal{A}}_K \rvert}{\text{vol}(\mathcal{A}_k)}    
    \end{align}
    
    through the generalised eigenvalue problem of symmetrically normalised graph Laplacian matrix \citep{Shi_Malik}, where $\mathcal{A}_k \quad \forall k \in [K]$ is a set of nodes in a graph which belongs to cluster $k$, and $\bar{\mathcal{A}}_k$ and $\text{vol}(\mathcal{A}_k)$ denote the complement and total sum of node degree of $\mathcal{A}_k$ respectively. By taking a normalisation with $\text{vol}(\mathcal{A}_k)$, rural bike stations with a small total node degree would be punished for clustering. For the KNN graph with $k$ nearest neighbours, the graph Laplacian $\mathbf{L}_{\mathrm{KNN}}^{(k)}$ and the symmetric normalised Laplacian $\tilde{\mathbf{L}}_{\mathrm{KNN}}^{(k)}$ are defined as 
    \begin{align*}
        \tilde{\mathbf{L}}_{\mathrm{KNN}}^{(k)} &= {\mathbf{D}_{\mathrm{KNN}}^{(k)}}^{-\frac{1}{2}} \mathbf{L}_{\mathrm{KNN}}^{(k)}{\mathbf{D}_{\mathrm{KNN}}^{(k)}}^{-\frac{1}{2}}\\
        \mathbf{L}_{\mathrm{KNN}}^{(k)} &= \mathbf{D}_{\mathrm{KNN}}^{(k)} - \mathbf{A}_{\mathrm{KNN}}^{(k)}
    \end{align*}
    where $\mathbf{D}_{\mathrm{KNN}}^{(k)}$ is the degree matrix of KNN graph with $k$ nearest neighbours. The symmetrically normalised graph Laplacian matrix is then utilised for obtaining a cluster assignment matrix $\mathbf{S} \in \mathbb{R}^{\lvert \mathcal{V} \rvert \times K}$ as follows \citep{Hamilton}: 
    \begin{enumerate}
        \item Calculate the symmetrically normalised Laplacian $\tilde{\mathbf{L}}_{\mathrm{KNN}}^{(k)}$ and obtain its second to $(K+1)$-th smallest eigenvalues and their corresponding eigenvectors;
        \item Form a matrix $\textbf{U}\in \mathbb{R}^{|\mathcal{V}|\times K}$ where columns are $K$ eigenvectors of $\tilde{\mathbf{L}}_{\mathrm{KNN}}^{(k)}$ from step 1;
        \item Apply $k$-means clustering across $\mathbf{U}_{[u]} \in \mathbb{R}^{K} \; \forall u\in \mathcal{V}$ and obtain $\mathbf{s}_u \in \mathbb{R}^{K}$, where $\mathbf{U}_{[u]}$ is the $u$-th row of \textbf{U} which corresponds to the graph node \textit{u};
        \item Form a cluster assignment matrix $\mathbf{S} \in \mathbb{R}^{\lvert \mathcal{V} \rvert \times K}$, where $u$th row corresponds to $\mathbf{s}_u$ obtained in the step 3.
    \end{enumerate}

    \subsubsection{Graph Coarsening}
    

    Given a node embedding matrix $\mathbf{Z} \in \mathbb{R}^{\lvert \mathcal{V} \rvert \times \ell}$ with $\ell$ latent variables and symmetrically normalised adjacency matrix of KNN graph $\tilde{\mathbf{A}}_{\mathrm{KNN}}^{(k)}={\mathbf{D}_{\mathrm{KNN}}^{(k)}}^{-\frac{1}{2}} \mathbf{A}_{\mathrm{KNN}}^{(k)} {\mathbf{D}_{\mathrm{KNN}}^{(k)}}^{-\frac{1}{2}} \in \mathbb{R}^{\textit{N}\times\textit{N}}$, the pooled node embedding and adjacency matrices are computed, respectively, as
    \begin{align*}
    \mathbf{Z}^{\mathrm{pool}}&=\mathbf{S}^\top\textbf{Z}\\  \mathbf{A}^{\mathrm{pool}}&=\mathbf{S}^\top\tilde{\mathbf{A}}_{\mathrm{KNN}}^{(k)}\mathbf{S}
    \end{align*}
  The entry $z_{i,j}^{\mathrm{pool}}$ in $\mathbf{Z}^{\mathrm{pool}} \in \mathbb{R}^{K \times \ell}$ denotes the pooled components of node embedding $j$ of nodes in cluster $i$. $\textbf{A}^{\mathrm{pool}}\in\mathbb{R}^{K\times K}$ is a symmetric matrix, with entry $a_{i,j}^{\mathrm{pool}}$ indicating the weighted sum of edges between clusters $i$ and $j$, and entries $a_{i,i}^{\mathrm{pool}}$ indicating the weighted sum of edges between the nodes in cluster $i$ \citep{bianchi2020spectral}.
    
\section{Proposed Solution}
    \subsection{Dataset}
    \subsubsection{Node Features and Weighted Edges}
\begin{figure}[!h]
        \centering
        \includegraphics[height=5.2cm,keepaspectratio]{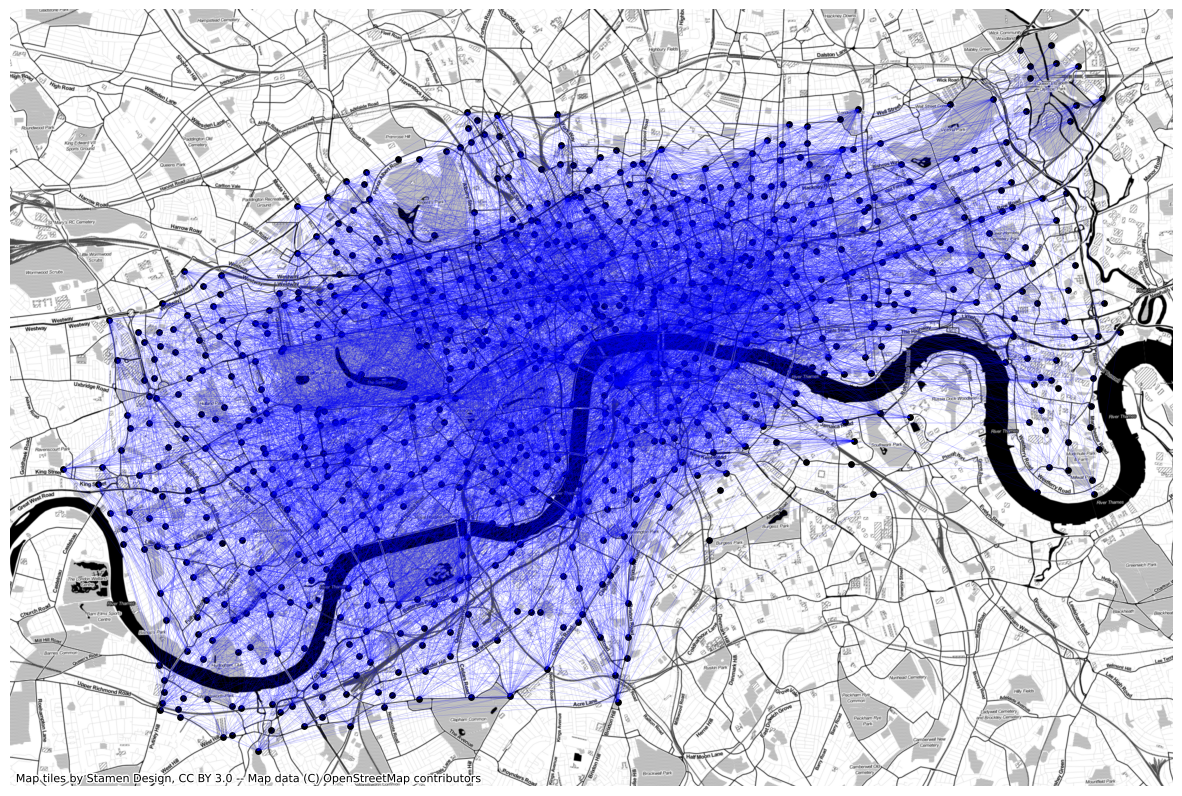}
        \includegraphics[height=5.2cm,keepaspectratio]{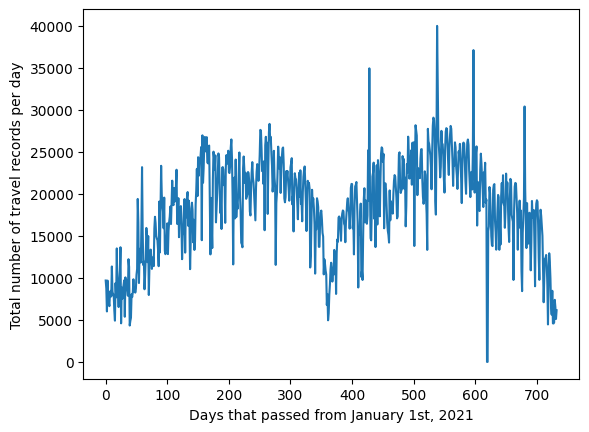}
        \caption{Visualised undirected graph of travel records from Santander Cycles on April 29th, 2021 (left), and the total number of travel records per day (right).} 
        \label{fig:travel_records}
    \end{figure}

    We first obtained the dataset of bike stations and daily travel records using the Santander Cycles API \citep{tfl} for the years 2021 and 2022, totalling 730 days. 798 stations were available as nodes and 13,483,777 travel records were available as edges. As seen in Figure \ref{fig:travel_records} on the right side, there is a seasonal pattern of demands for the service, with a few outliers. To avoid confounded prediction, 7 samples which had more than 30,000 or less than 5,000 travel records were omitted. Daily travel records were then aggregated by the same origin and destination pair, yielding 723 $(=730-7)$ undirected graphs. We then mapped node features of 6 land-use characteristics \citep{fleischmann_arribas-bel_2022} and the number of households in the smallest area unit by 10 family structures \citep{census_2021_results_2022} onto the nodes of bike stations.

   \subsubsection{Graph Features and Target Labels} \label{graphcoars}
    For the graph features and target labels, the meteorological dataset the \citep{ecad} observed daily at Heathrow was used. This can be considered as approximation to London's weather conditions. To capture seasonal patterns as seen in Figure \ref{fig:travel_records}, we consider the mean, maximum, and minimum daily temperatures after normalising to 0 mean and 1 standard deviation, as graph features. For the target labels, we created three categories of weather: (i) `rainy', if precipitation is more than 0mm; (ii) `cloudy', if daily cloud coverage is more than 20\%; and otherwise (iii) `sunny'. This is combined with a binary indicator of weekends or holidays, obtained from the Bank Holidays API \citep{bank_holidays_api}, yielding in total 6 categories. A summary is presented in Table \ref{tab:features_description}.

    \begin{table}[!h]
        \centering
        \small
        \begin{tabular}[t]{cccc}
            Name & Category & Data Type & Dimensions \\
            \hline
            Land-use Characteristics & Node Feature & Categorical & 6\\
            Number of Households by Demographics & Node Feature & Continuous & 10\\
            Temperatures & Graph Feature & Continuous & 3\\
            Weather $\times$ (Weekday or Weekend/Holiday) & Target Label & Categorical & $3 \times 2 = 6$\\
            \hline
        \end{tabular}
        \caption{Summary of node features, graph features, and target labels.}
        \label{tab:features_description}
    \end{table}
    
    \subsection{Models}
    \subsubsection{Model Architectures}
    
\begin{figure}[!h]
        \centering
        \includegraphics[width=\linewidth]{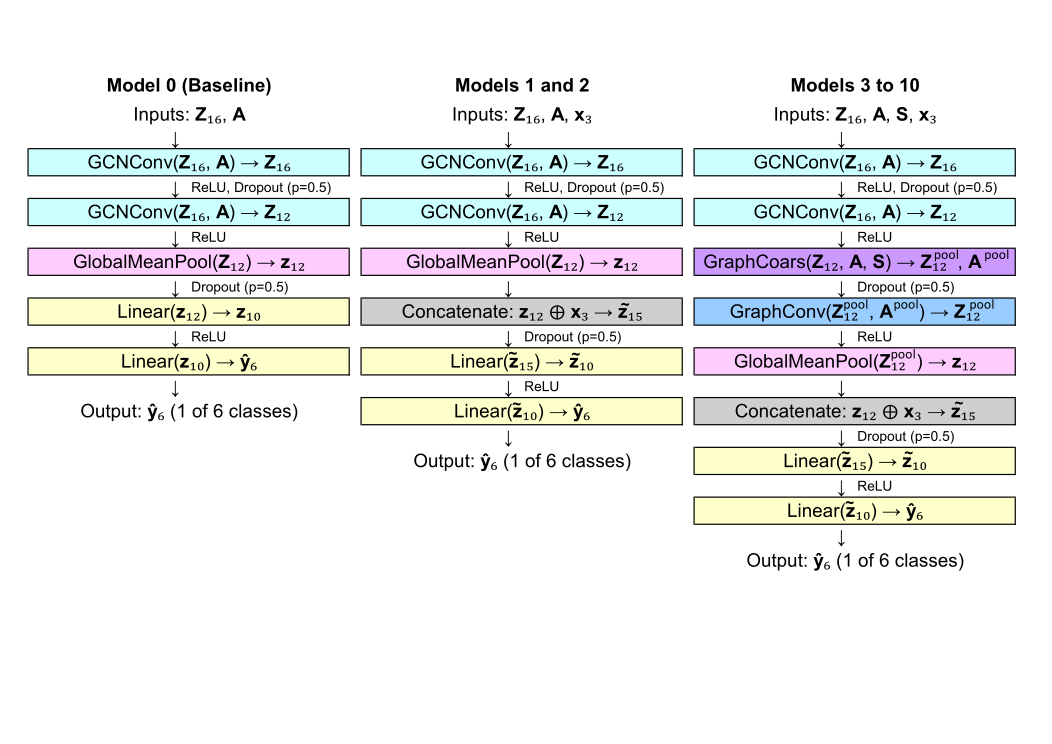}
        \caption{A simplified diagram of the models. Each layer is written as $\mathrm{Operator}(\mathrm{Input}) \rightarrow \mathrm{Output}$. Subscripts denote the dimension (number of features).} 
        \label{models}
    \end{figure}

    A total of 11 models are evaluated in this study. The baseline model (Model 0) uses 2 $\mathrm{GCNConv}$ layers \citep{kipf2017semisupervised}, a global mean pooling layer as described in Section \ref{globalmeanpool} and 2 Linear (or Fully-Connected) layers, with rectified linear unit (ReLU) activation function $\mathrm{ReLU}(x) = \max\{0, x\}$ and dropout layers (with probability $p=0.5$) in between. All other models also consider the mean, minimum and maximum daily temperatures as input features. Models 3 to 10 additionally employ the $\mathrm{GraphCoars}$ operator as described in Section \ref{graphcoars}, followed by a $\mathrm{GraphConv}$ layer \citep{morris2021weisfeiler}. The construction of $\mathbf{A}$ and $\mathbf{S}$ are different for each model and is summarised in Table \ref{tab:model_notation}. A simplified diagram of the models with condensed notation is presented in Figure \ref{models}.

    
        \begin{table}[!h]
        \centering
        \small
        \begin{tabular}[t]{c|cc}
            Model & $\mathbf{A}$ & $\mathbf{S}$ with 32 clusters\\
            \hline 
            0 & unweighted & n/a\\
            1 & unweighted & n/a\\
            2 & weighted & n/a \\
            3 & unweighted & unweighted $\mathbf{A}$ \\
            4 & unweighted & KNN ($k=5$) \\
            5 & unweighted & KNN ($k=50$) \\
            6 & unweighted & KNN ($k=150$) \\
            7 & weighted & unweighted $\mathbf{A}$ \\
            8 & weighted & KNN ($k=5$) \\
            9 & weighted & KNN ($k=50$) \\
            10 & weighted & KNN ($k=150$) \\
            \hline
        \end{tabular}
        \caption{Model specific construction of $\mathbf{A}$ and $\mathbf{S}$.}
        \label{tab:model_notation}
    \end{table}
        
    \subsubsection{Loss Function}
        

    \begin{align} \label{eq:ce}
        \mathcal{L} = \sum_{u \in \mathcal{V}_{\text{train}}}-\log(\text{softmax}(\hat{\mathbf{y}}_{\mathcal{G}}, \mathbf{y}_{\mathcal{G}}))
    \end{align}
    All models were trained with the same loss function of the cross-entropy loss shown in Equation \eqref{eq:ce}. Cross-entropy loss measures the performance of the classification model where the prediction vector $\hat{\mathbf{y}}_{\mathcal{G}}$ forms a probability simplex. Cross-entropy loss increases as $\hat{\mathbf{y}}_{\mathcal{G}}$ diverges from $\mathbf{y}_{\mathcal{G}}$, hence we want to minimise this value. A perfect classifer has a cross-entropy loss of 0.
    
    \subsubsection{Computational complexity}
    Most of the graph neural networks suffer from the issue of time and space complexity. Graph convolution methods develop schemes for filtering a neighbourhood of a graph which are created on the top of the adjacency matrix.  Essentially, a graph classification takes an adjacency matrix of size $\mathcal{O}({\lvert \mathcal{V} \rvert}^2)$ as input. As the graph size increases, i.e., the number of graph nodes $\lvert \mathcal{V} \rvert$, increases, the graph convolution methods require the quadratic graph size for each layer \citep{time_complexity}. Moreover, a $L$-layer GCN model has time complexity $\mathcal{O}(L \cdot \lvert \mathcal{V} \rvert \cdot d^2)$ and space complexity $\mathcal{O}(L \cdot \lvert \mathcal{V} \rvert \cdot d +L\cdot d^2)$, where $d$ is the dimension (or number of channels).
When compared to Models 0 to 2, the computational time for Models 3 to 11 is significantly larger, due to the added step of graph coarsening using spectral clustering. For the spectral clustering algorithm, the most computationally expensive step is the computation of the eigenvalues and eigenvectors for the Laplacian matrix. Given ${\lvert \mathcal{V} \rvert}$ data points, the algorithm forms an adjacency matrix $\mathbf{A} \in \mathbb{R}^{\lvert \mathcal{V} \rvert \times \lvert \mathcal{V} \rvert}$ and computes the eigenvectors for this matrix. This process has an overall time complexity of $\mathcal{O}({\lvert \mathcal{V} \rvert}^3)$ \citep{sctime}. Thus, we can state that the time complexity of spectral clustering-based graph coarsening is proportional to the time complexity of spectral clustering.

    \section{Numerical Experiments}
    \subsection{Expected Results}
    Table \ref{tab:count_data} shows the count for the weekend/holiday or weekday in the dataset used. It is evident that the dataset is imbalanced in terms of the count, and thus could potentially create some bias in the predictions, as the models could overfit and tend to predict the majority classes (Rainy Weekday or Cloudy Weekday) as the weather along with day type. With the baseline model, we expect the accuracy to be at least 35\%. Another benchmark that can be set for the performance measures is the percentage of the majority class in the dataset. Cloudy weekday comprises of 32\% of the dataset and we expect the measures to be higher than that. 
    \begin{table}[!h]
        \centering
        \small
        \begin{tabular}{c|ccc}
          Weekend/Holiday   & Rainy & Sunny & Cloudy \\
          \hline
            0 & 222 & 50 & 226\\
            1 & 98 & 24 & 103\\
            \hline
        \end{tabular}
        \caption{Count of Weather Type in the dataset}
        \label{tab:count_data}
    \end{table}

    \subsection{Experimental Results}    
    All models were trained for 500 epochs with the same configuration: the data is split into 80\% for training and 20\% for validation/testing, with a batch size of 32. The Adam optimiser with learning rate $\gamma=0.001$ and weight decay $\lambda=0.0001$ (with $L_2$ regularisation)  were used. To evaluate the performance of the models, we report the Accuracy = $\frac{1}{N} \sum_{i=1}^N \mathbf{1}{\{y_i = \hat{y}_i\}} $ for each of the models.
    
    
\begin{table}[!h]
    \centering
    \small
    \begin{tabular}{c|cc|cc}
         & Training & Validation & Training & Validation\\
        Model & Loss & Loss & Accuracy & Accuracy\\
        \hline
        0 & 1.47 $\pm$ 0.02 & 1.49 $\pm$ 0.04 & 0.37 $\pm$ 0.01 & 0.36 $\pm$ 0.01\\
        1 & 1.28 $\pm$ 0.03 & 1.40 $\pm$ 0.07 & 0.43 $\pm$ 0.02 & 0.41 $\pm$ 0.06\\
        2 & 1.32 $\pm$ 0.03 & 1.44 $\pm$ 0.07 & 0.41 $\pm$ 0.02 & 0.41 $\pm$ 0.05\\
        3 & 1.24 $\pm$ 0.06 & 1.24 $\pm$ 0.11 & 0.45 $\pm$ 0.02 & 0.43 $\pm$ 0.04\\
        4 & 1.20 $\pm$ 0.04 & \textbf{1.14 $\pm$ 0.06} & \textbf{0.46 $\pm$ 0.02} & \textbf{0.48 $\pm$ 0.03}\\
        5 & 1.27 $\pm$ 0.04 & 1.30 $\pm$ 0.05 & 0.45 $\pm$ 0.02 & 0.41 $\pm$ 0.02\\
        6 & \textbf{1.19 $\pm$ 0.04}	& 1.22 $\pm$ 0.09 & 0.46 $\pm$ 0.02 & 0.45 $\pm$ 0.04\\
        7 & 1.39 $\pm$ 0.02 & 1.62 $\pm$ 0.03 & 0.40 $\pm$ 0.02 & 0.35 $\pm$ 0.01\\
        8 & 1.36 $\pm$ 0.03 & 1.54 $\pm$ 0.04 & 0.41 $\pm$ 0.02 & 0.39 $\pm$ 0.02\\
        9 & 1.36 $\pm$ 0.03 & 1.60 $\pm$ 0.04 & 0.41 $\pm$ 0.02 & 0.37 $\pm$ 0.02\\
       10 & 1.36 $\pm$ 0.02 & 1.57 $\pm$ 0.04 & 0.41 $\pm$ 0.02 & 0.38 $\pm$ 0.02\\
        \hline
    \end{tabular}
    \caption{Summary of results from all models where model numbers correspond to those on Table \ref{tab:model_notation}. (\textit{mean} $\pm$ \textit{standard deviation}) between 400 and 500 epochs are shown in each cell. The best \textit{mean} score in each column is highlighted in bold.}
    \label{tab:results}
\end{table}

Table \ref{tab:results} as a result presents several key insights about the performance of the different models. Firstly, when compared to the baseline model (Model 0), Models 1 and 2 showed improvement in both the cross-entropy and accuracy, due to the concatenation of graph features. This could be interpreted that the mean, maximum, and minimum temperatures of London could capture the pattern of seasonal patterns in demands for bikes, and additional features on graph level could help the prediction in general through the concatenation operation. Secondly, the implementation of graph coarsening further improved the models, among which Model 4 (GCN with graph coarsening based on a KNN graph with $k=5$) scored the best with validation loss of 1.14 and accuracy of 0.48. Model 4 also outperformed Model 3 with validation loss of 1.24 and accuracy of 0.43, implying that the explicit capture of geographical contiguity succeeded in the efficient aggregation of node embeddings, as \textit{Spatial Graph Coasening}. Furthermore, when comparing graphs with weighted and unweighted edges, graph coarsening had a greater impact on weighted graphs. While Models 7 to 10 did not outperform Model 2 in terms of the mean validation loss and accuracy, their standard deviations decreased (i.e. 0.05 of Model 2 $\to$ 0.02 of Models 8 to 10 for validation accuracy), implying the stabilising effect of graph coarsening while training the models. Finally, the best-performing models were achieved with KNN graphs with $k=5$, for both weighted and unweighted edges. This suggests that increasing the complexity of KNN graphs do not necessarily improve the efficiency of aggregating node embeddings in neighbours. One of the possible reasons behind this could be that, as $k$ increases the graph would be closer to a complete graph, where no clusters could be obtained through Spectral Clustering.

\section{Conclusion}
This study compared 11 different models to determine which performs the best for predicting weather and day type on the basis of bike-sharing demand. We introduced two new architectures: the (i) concatenation operator of graph features; and (ii) spatial graph coarsening based on geographical contiguity. As expected, the model that implemented spatial graph coarsening in the training phase with 5-nearest neighbours performed the best. However, the best model could only reach up to the accuracy of 48\%. This could be potentially because of certain limitations of the models. We only considered data for the years 2021 and 2022, totaling 723 days, which could have added some bias in the predictions as the dataset is relatively small. Given more resources, we could have deployed a larger dataset which would have led to better training and accuracy. Moreover, for the simplicity of predictions, we considered a cross-product of day type and weather type. With a reasonable amount of data, we could have created multiple categories with a combined loss function for the multi-label classification task. Also, the weather data considered was an approximation of London's weather. If a more granular dataset would have been available, such as hourly weather conditions differentiated by district, we could have split the travel records into finer time and geographic units. Lastly, we only used GCN for the message-passing framework. Given more time, we could have used alternative frameworks like GraphSAGE and Graph Attention Networks (GAT) \citep{Hamilton} which could potentially improve the models. Accuracy may not be a proper measure for imbalanced datasets since it does not distinguish between the numbers of correctly classified examples of different classes and we could have used other metrics like $F_1$-score as it considers false positives and false negatives. However, the framework that we provide can be used as a base for further work and research. Lastly, the cluster size was set to 32 for the comparison between different spectral clustering. The choice of batch size makes a trade-off between the stability of the model and computational complexity. After experimenting it was found that smaller batch sizes tend to easily overfit the data. 
 
\bibliography{references}
    
\end{document}